\RequirePackage{fix-cm}
\documentclass[twocolumn]{svjour3}          
\smartqed  
\usepackage{graphicx}
\usepackage{mwe}
\usepackage{subcaption}
\usepackage{amsmath}
\usepackage[utf8x]{inputenc}
\usepackage{lscape}
\usepackage{afterpage}
\usepackage{longtable}
\usepackage{array,ragged2e}
\usepackage{color, colortbl}
\definecolor{Gray}{gray}{0.8}
\definecolor{LightGray}{gray}{0.9}
\usepackage[numbers,square]{natbib}
\usepackage{svg}

\begin{document}
\sloppy

\title{Facial emotion expressions in human-robot interaction: \\ 
A survey
}


\author{Niyati Rawal         \and
        Ruth Maria Stock-Homburg 
}


\institute{Niyati Rawal, M.Eng. \at
              Technical University Darmstadt \\
              Hochschulstr. 1 \\
              64295 Darmstadt, Germany \\
              Phone: +496151162466\\
              \email{niyati.rawal@tu-darmstadt.de}           
           \and
           Ruth Maria Stock-Homburg, PhD, PhD \at
           Professor for Marketing HR \\
           \email{RSH@tu-darmstadt.de} \\
           (Corresponding author)
}

\date{Received: date / Accepted: date}

\maketitle

\begin{abstract}
Facial expressions are an ideal means of communicating one's emotions or intentions to others. This overview will focus on human facial expression recognition as well as robotic facial expression generation. In the case of human facial expression recognition, both facial expression recognition on predefined datasets as well as in real-time will be covered. For robotic facial expression generation, hand-coded and automated methods i.e., facial expressions of a robot are generated by moving the features (eyes, mouth) of the robot by hand-coding or automatically using machine learning techniques, will also be covered. There are already plenty of studies that achieve high accuracy for emotion expression recognition on predefined datasets, but the accuracy for facial expression recognition in real-time is comparatively lower. In the case of expression generation in robots, while most of the robots are capable of making basic facial expressions, there are not many studies that enable robots to do so automatically. In this overview, state-of-the-art research in facial emotion expressions during human-robot interaction has been discussed leading to several possible directions for future research. 
\end{abstract}

\section{Introduction}
\label{intro}
Robots are no longer just machines being used in factories and industries. There is a growing need and demand towards robots sharing space with humans as collaborative robotics or assistive robotics \cite{Kirgis2016,5751968}. Robots are, now, increasingly being deployed in a variety of domains as receptionists \cite{Stock2016EmotionTF}, educational tutors \cite{hoffman-zuckerman,kanda}, household supporters~\cite{saerbeck} and caretakers \cite{dautenhahn,hoffman-zuckerman,kozima,tanaka}. Thus, there is a need for these social robots to effectively interact with humans, both verbally and non-verbally. Facial expressions are non-verbal signals that can be used to indicate one’s current status in a conversation, e.g., via backchanneling or rapport \cite{Bavelas2011,Drolet2000RapportIC}.

\begin{figure*}[ht]
\includegraphics[width=\textwidth]{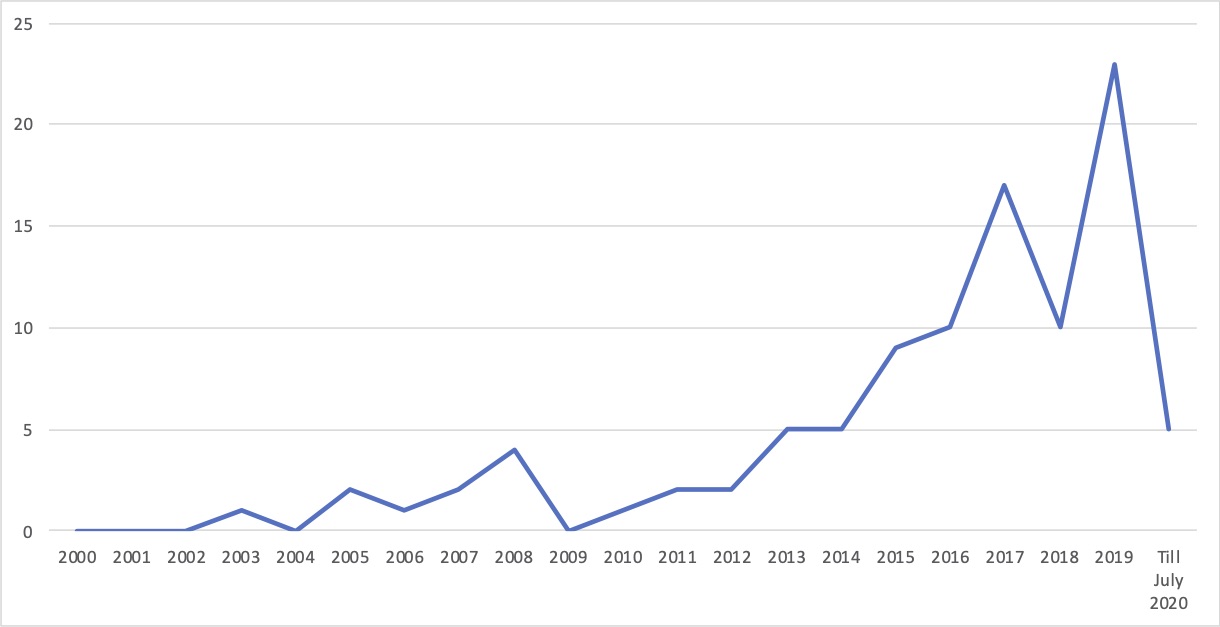}
\caption{Publications on emotion recognition of human faces during HRI and generation of facial expressions of robots}
\label{fig:pub}
\includegraphics[width=\textwidth]{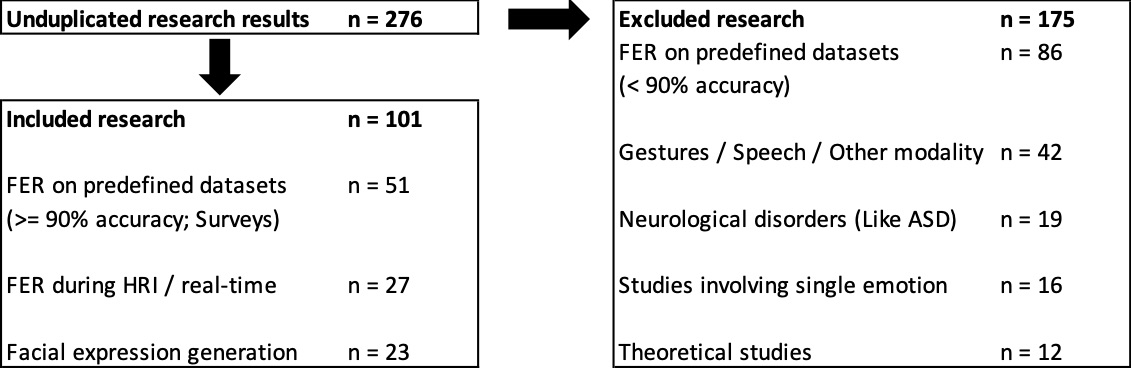}
\caption{Flowchart of the literature screening process}
\label{fig:screen}
\end{figure*}

Perceived sociability is an important aspect in human-robot interaction (HRI) and users want robots to behave in a friendly and emotionally intelligent manner \cite{Graaf2016,Ray2008,952716,Hoffman2006}. For social robots to be more anthropomorphic and for human-robot interaction to be more like human-human interaction (HHI), robots need to be able to understand human emotions and appropriately respond to those human emotions. Stock and Merkle show that emotional expressions of anthropomorphic robots become increasingly important in business settings as well \cite{7917585,stock2018can}. The authors of \cite{stock2019robotic} emphasize that robotic emotions are particularly important for the acceptance of a robot by the user. Thus, emotions are pivotal for HRI \cite{stockhomburg2021}. In any interaction, 7\% of the affective information is conveyed through words, 38\% is conveyed through tone, and 55\% is conveyed through facial expressions~\cite{mehrabian}. This makes facial expressions an indispensable mode of affective communication. Accordingly, numerous studies have examined facial expressions of emotions during HRI \cite[e.g.][]{cid2014,meghdari2016,7860248,Bera2019TheEI,liu2017,8630711,7363421,ruiz-garcia2018,churamani-barros,6696662,9071837,ge2008,horii,liu2017,meghdari2016}. 

In any HHI, human beings first infer the emotional state of the other person and then accordingly generate facial expressions in response to their peer. The generated emotion could be a result of parallel empathy (generating the same emotion as the peer) or reactive empathy (generating emotion in response to the peer's emotion) \cite{Davis2018}. Similarly, in the case of HRI, we would like to study robots recognizing human emotion as well as robots generating their emotion as a response to human emotion. 

There has been a growth in the number of papers on facial expressions in HRI in the last decade. Between 2000 and 2020 (see Figure \ref{fig:pub}), there has been a gradual increase in the number of publications. Thus, the overarching research question is: What has been done so far on facial emotion expressions in human-robot interaction, and what still needs to be done?

\begin{figure*}[t]
\includegraphics[width=\textwidth]{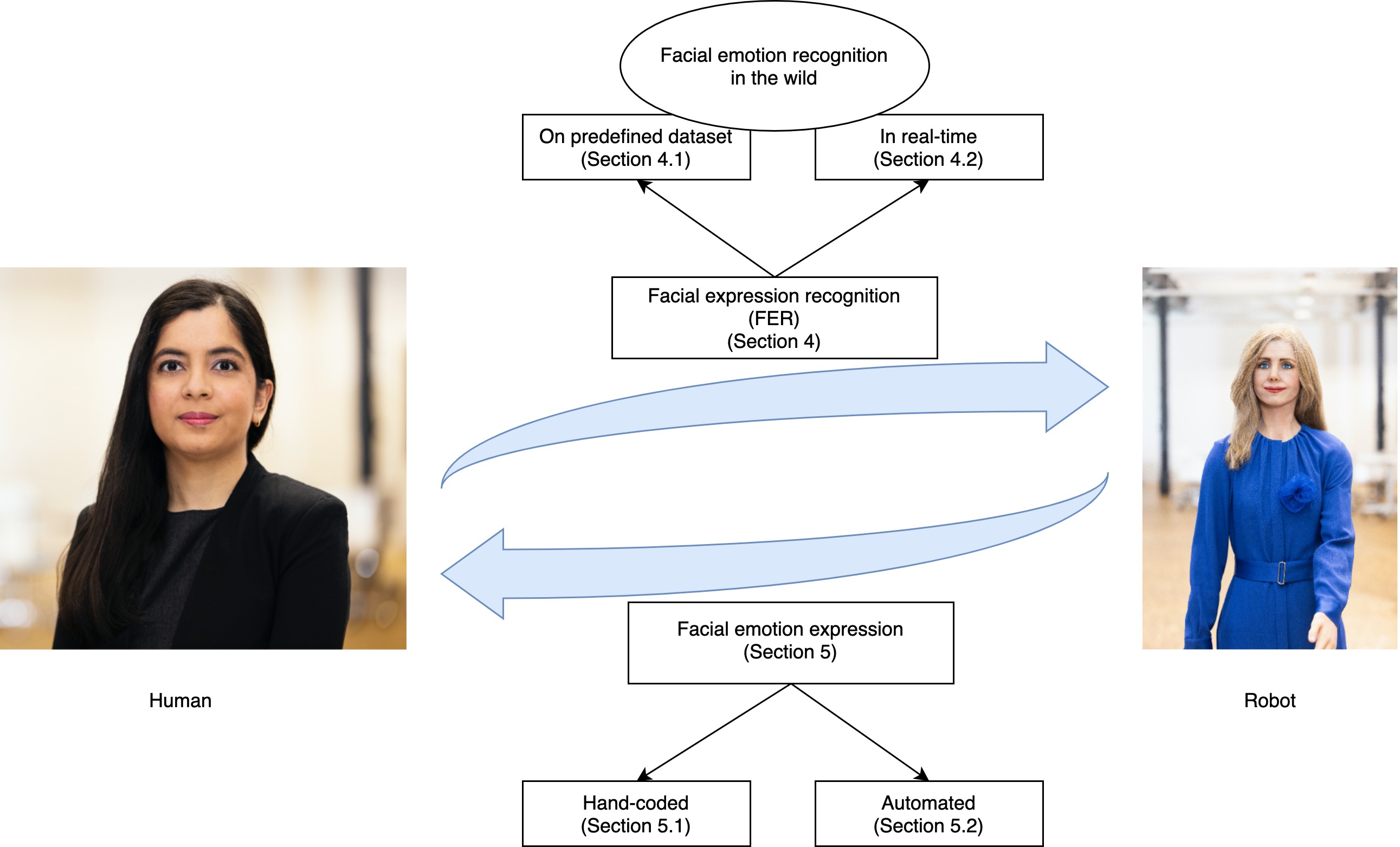}
\caption{Framework of the overview}
\label{fig:framework}
\end{figure*}

In section \ref{sec:1} the framework of the overview is outlined, followed by the method of selection of studies in section \ref{sec:2}. Recognition of human facial expressions and generation of facial expressions by robots are covered in sections \ref{sec:3} and \ref{sec:4}. The current state of the art and future research are discussed in section \ref{sec:5} with the conclusion in section \ref{sec:6}.

\section{Framework of the overview}
\label{sec:1}
This overview focuses on two aspects: (1) recognition of human facial expressions and (2) generation of facial expressions by robots. The review framework (Figure \ref{fig:framework}) is based on these two streams. (1) Recognition of human facial expressions is further subdivided depending on whether the recognition takes place on (a) a predefined dataset or in (b) real-time. (2) Generation of facial expressions by robots is also subdivided depending on whether the facial generation is (a) hand-coded or (b) automated, i.e., facial expressions of a robot are generated by moving the features (eyes, mouth) of the robot by hand-coding or automatically using machine learning techniques.

\section{Method}
\label{sec:2}
Studies with the keywords “facial expression recognition AND human-robot interaction / HRI", "facial expression recognition" and "facial expression generation AND human-robot interaction / HRI" between 2000 and 2020 were reviewed on Google Scholar.

In this overview, studies that use voice or body gestures as a modality for emotional expression but do not involve facial expressions are not included. Studies that involve HRI with humans having mental disorders like autism are also not included. Furthermore, studies that work on single emotion such as recognition of smile or facial expression generation of anger are not included. In total, 175 studies of 276 were rejected (Figure \ref{fig:screen}).

In Table \ref{tab0}, various studies on facial expression recognition are listed. Here, studies with an accuracy of greater than 90\% for facial expression recognition on predefined datasets are selected. 

For real-time facial expression recognition, all studies that perform facial expression recognition in a human-robot interaction scenario are listed.

\begin{figure*}[t]
\includegraphics[width=\textwidth]{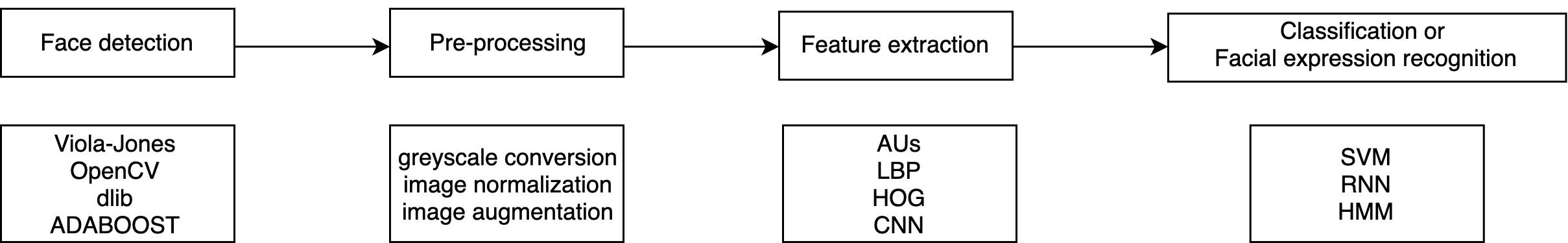}
\emph{Notes}: Action Units (AUs), Local Binary Patterns (LBP), Histogram of Oriented Gradients (HOG), Convolutional Neural Network (CNN), Support Vector Machine (SVM), Recurrent Neural Network (RNN), Hidden Markov Model (HMM)
\caption{Process of facial expression recognition in machine learning (adapted from Canedo and Neves \cite{canedo})}
\label{fig:fig3} 
\end{figure*}

\begin{figure*}[t]
\includegraphics[width=\textwidth]{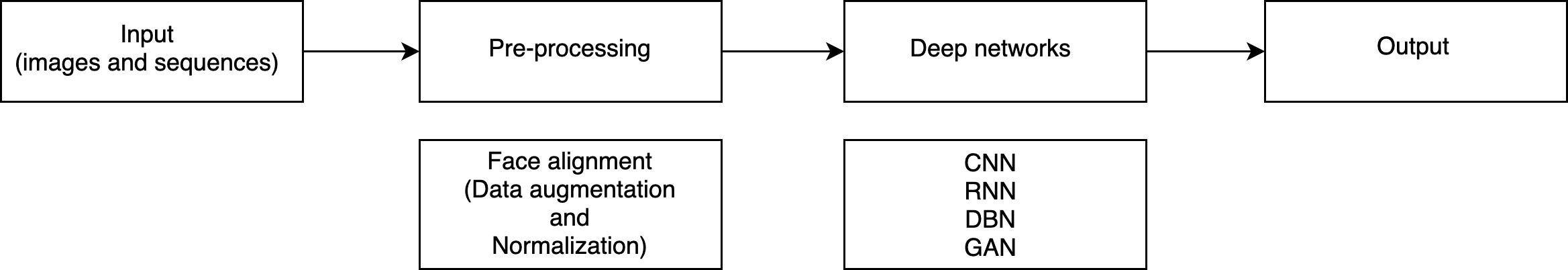}
\emph{Notes}: Convolutional Neural Network (CNN), Recurrent Neural Network (RNN), Deep Belief Network (DBN), Generative Adversarial Network (GAN)
\caption{Process of facial expression recognition in deep leaning (adapted from Li and Deng \cite{9039580})}
\label{fig:deeplearn}
\end{figure*}

\begin{table*}[!htp]
{
\caption{Details about the commonly used algorithms}
\label{tab4}
\begin{tabular}{p{1.4cm} p{3.15cm} p{3.15cm} p{3.15cm} p{3.15cm}p{1.4cm}}
\hline
\textbf{Algorithm (year)} & \textbf{Major Purpose} & \textbf{Application areas} & \textbf{Advantages} & \textbf{Disadvantages} & \textbf{Frequency of use}\\
\hline
KNN (1950s) & Storing of all available cases and classifies new instances by measuring the similarity or difference between two instances using a distance function \cite{4406010} & Classification of facial expressions & No training required before making predictions & Great computational complexity, because the distance between every sample should be calculated in order to classify \cite{4667244} & 1\\
\hline
HMM (1960s) & Given a sequence of observations, decode the hidden states \cite{18626} & Perform FER based on dynamic data input i.e. videos & Capture the dependencies between consecutive measurements & Information from states in the preceding time steps (not the previous time step) cannot be captured & 1\\
\hline
RNN (1980s) & Learn temporal dependencies \cite{lipton2015critical} & Perform FER based on dynamic data input i.e. videos & Useful in modelling sequential data & Difficult to learn long-term temporal dependencies as gradients explode or vanish over many time steps \cite{bengio1994learning} & - \\
\hline
CNN (1989) & Given a set of images, extracts featural information such as edges and performs the classification task \cite{5537907} & Learn the spatial features in an image i.e. perform FER based on static data input & Easier to train and generalizes much better than networks with full connectivity between adjacent layers \cite{lecun2015deep} & Requires a lot of data to prevent over-fitting \cite{yamashita2018convolutional} & 13\\
\hline
SVM (1995) & A supervised learning algorithm that can be used for classification or regression on small datasets \cite{788640} & Classification of facial expressions & Robust to over-fitting \cite{taira1999feature}  & Does not perform well with large training samples \cite{liu2017svm} & 3 \\
\hline
LSTM (1997) & Learn long-term temporal dependencies \cite{gers2000recurrent} & Perform FER based on dynamic data input i.e. videos & Ability to learn long-term temporal dependencies \cite{hochreiter1997long} & Computationally expensive to train & 1\\
\hline

\end{tabular}
}
\emph{Notes}: k-nearest neighbor (KNN), hidden markov model (HMM), recurrent neural network (RNN), convolutional neural network (CNN), support vector machine (SVM) and long short-term memory (LSTM)
\end{table*}

\section{Recognition of human facial expressions}
\label{sec:3}

Earlier, facial expression recognition (FER) consisted of the following steps: detection of face, image pre-processing, extraction of important features and classification of expression (Figure \ref{fig:fig3}). As deep learning algorithms have become popular, the pre-processed image is directly fed into deep networks (like CNN, RNN etc.) to predict an output \cite{9039580} (Figure \ref{fig:deeplearn}).

\onecolumn
\afterpage{
\begin{landscape}
\begin{center}
\newcolumntype{P}[1]{>{\RaggedRight\arraybackslash}p{##1}}
\begin{longtable}{|P{2cm}|P{2cm}|P{3cm}|P{1.5cm}|P{3.5cm}|P{9cm}|}
\caption{Detailed information about studies on emotion recognition and Human Robot Interaction (HRI)} \label{tab6} \\

\hline \multicolumn{1}{|P{2cm}|}{\textbf{Authors (year)}} & \multicolumn{1}{P{2cm}|}{\textbf{Recognition mode}} &
\multicolumn{1}{P{3cm}|}{\textbf{Recognized emotions}} &
\multicolumn{1}{P{1.5cm}|}{\textbf{Robot}} &
\multicolumn{1}{P{3.5cm}|}{\textbf{Algorithm type}} &
\multicolumn{1}{P{9cm}|}{\textbf{Major findings}} \\ \hline 
\endfirsthead

\multicolumn{5}{c}%
{{}} \\
\hline \multicolumn{1}{|P{2cm}|}{\textbf{Authors (year)}} & \multicolumn{1}{P{2cm}|}{\textbf{Recognition mode}} &
\multicolumn{1}{P{3cm}|}{\textbf{Recognized emotions}} &
\multicolumn{1}{P{1.5cm}|}{\textbf{Robot}} &
\multicolumn{1}{P{3.5cm}|}{\textbf{Algorithm type}} &
\multicolumn{1}{P{9cm}|}{\textbf{Major findings}} \\ \hline
\endhead

\endfoot

\hline 
\endlastfoot
Ahmed et al. (2019) \cite{8858529} & face & angry, disgust, fear, happy, neutral, sad, surprise & - & CNN with data augmentation & The model achieved an accuracy of more than 90\% for each emotion as it could classify geometrically displaced facial images. \\
\hline
Barros et al. (2015) \cite{7363421} & face & positive, negative, neutral & iCub & CNN & The network is able to recognize emotions from different  environments,   different  subjects  per-forming spontaneous expressions, and in real-time. \\
\hline
Bera et al. (2019) \cite{Bera2019TheEI} & face & happy, sad, angry, neutral & Pepper & Bayesian inference and CNN & A multi-channel model to classify pedestrian features into four categories of emotions. Emotional detection accuracy of 85.33\% was observed in the validation results. \\
\hline
Byeon and Kwak (2014) \cite{byeon2014facial} & face & happiness, sadness, anger, surprise, disgust, fear & - & 3D-CNN & The experimental results on video-based facial expression database revealed that the method showed a good performance in comparison to the conventional methods such as PCA and TMPCA. \\
\hline
Chen et al. (2018) \cite{8630711} & face & angry, disgust, fear, happy, sad, surprise, neutral & XiaoBao & CNN (VGG-16) & The facial expression recognition in the wild (FERW) model can recognize facial expressions in the real-world with an accuracy of 79\% and a real-time positive emotion incentive system (PEIS) was able to enhance user experience. \\
\hline
Cid et al. (2014) \cite{cid2014} & face, voice & sad, happy, fear, anger, neutral & Muecas & Dynamic Bayesian Network & 
The Bayesian approach to the emotion recognition problem presents good results for real-time applications with untrained users in an uncontrolled environment. \\
\hline
Dand{\i}l and {\"O}zdemir (2019) \cite{dandil2019real} & face & anger, fear, happy, surprise, sad, neutral & - & CNN & Successful results obtained on real-time video, in changing light and environment conditions with 80\% accuracy. \\
\hline
Deng et al. (2017) \cite{deng2019cgan} & face & surprise, sad, neutral, happy, fear, disgust, anger & - & cGAN &
Can learn not only the regions related to expression but also maximally capturing nuanced characteristics relevant to expression and then transforming the original expression to another expression with identity and other factors preserved. \\
\hline
Faria et al. (2017) \cite{8172395} & face & afraid, angry, happy, neutral, disgusting, sad, surprised & Nao & Dynamic Bayesian Mixture Model (DBMM) & An overall accuracy around 85\% on KDEF dataset and 80\% on tests on-the-fly during human-robot interaction.\\
\hline
Ge et al. (2008) \cite{ge2008} & face & happy, sad, fear, disgust, anger, surprise & robotic head & SVM & The experimental results showed that the proposed nonlinear facial mass-spring model coupled with the SVM classifier is effective to recognize the facial expressions compared with the linear mass-spring model. \\
\hline
Inthiam et al. (2019) \cite{inthiam} & face and bodily movement & positive mood and negative mood & - & HMM & Facial expression alone may be misleading, since it may not be a true expression of inner feeling. By including bodily expression in the analysis, the estimation model gave a better result with estimation accuracy over 70\% \\
\hline
Li et al. (2019) \cite{8760246} & face & happiness, anger, disgust, fear, sadness, surprise & Harley & CNN-LSTM & CNN and LSTM are combined to exploit their advantages in the proposed model and the system is applied to a humanoid robot to demonstrate its practicability for improving the HRI.\\
\hline
Liu et al. (2017) \cite{liu2017} & face & happy, angry, surprise, fear, disgust, sad, and neutral & mobile robot & ELM classifier & Recognized human emotions with 80\% accuracy. \\
\hline 
Liu et al. (2019) \cite{8823409} & face & anger, scared, sad, surprise, neutral, disgust & - & CNN & An average weighting method was proposed to avoid potential errors in real-time facial expression recognition based on the traditional convolutional neural network. \\
\hline 
Lopez-Rincon (2019) \cite{8673111} & face & sadness, happiness, surprise, anger, disgust, fear & Nao & AFFDEX SDK and CNN & The global emotion classification accuracy of the CNN is better than the AFFDEX system, but the face detection in AFFDEX is slightly better. \\
\hline
Martin et al. (2008) \cite{4813412} & face & anger, disgust, fear, surprise, neutral, happy, sad & - & Active Appearance Model (AAM), Multi Layer Perceptron (MLP) and SVM & Compared three different facial expression classifiers (AAM classifier set, MLP and SVM). \\
\hline
Meghdari et al. (2016) \cite{meghdari2016} & face & happiness, sadness, fear, anger, surprise & Alice & ANN & Can recognize emotions with 92.52\% accuracy in real-time. \\
\hline
Nunes (2019) \cite{nunes} & face and upper body & anger, disgust, fear, happiness, sadness, surprise and neutral & - & CNN & Bimodal approach (86.6\% accuracy) based on emotion recognition from both face and upper body produced better results than monomodal approach.\\
\hline
Romero et al. (2013) \cite{romero2013novel} & face, voice, body gestures & neutral, fear, angry, sad, happy & - & Dynamic Bayesian Network & The standardized variables associated to the AUs of the user were obtained, allowing real-time recognition of each facial expression with different factors, such as lighting conditions, gender, unusual facial features (like injuries or scars), among others. \\
\hline
Ruiz-Garcia et al. (2018) \cite{ruiz-garcia2018} & face & surprise, happy, disgust, angry, fear, sad & - & CNN & Images from 28 participants were collected in an uncontrolled environment to test our CNN emotion recognition model, resulting in a classification rate of 73.55\%. \\
\hline
Shi et al. (2019) \cite{shi2019visual} & face and body & interested, distracted, confused & Pepper & k-Nearest Neighbour (KNN) &
A multi-student affect recognition system which, starting from eight basic emotions detected from facial expressions, can infer higher emotional states relevant to a learning context, such as “interested”, “distracted” and “confused”. \\
\hline
Simul et al. (2016) \cite{7860248} & face & neutral, surprise, sad, smile, angry & Ribo & SVM & Robot Ribo recognizes human facial expression, facial gesture movement and detects human gender in real-time. \\
\hline
Vithanawasam and Madhusanka (2019) \cite{vithanawasam19} & face and upper body & anger, fear, bored & - & Fisherface algorithm / position of arms & Anger was correctly predicted 81.54\% of the times, followed by bored (72.20\%) and fear (68.37\%). The results were sensitive to the lighting conditions. \\
\hline
Webb et al. (2020) \cite{9207494} & face & angry, disgust, fear, happy, neutral, sad, surprise, neutral & Nao & CNN & When evaluated on novel data with nonuniform conditions taken by a Nao robot an accuracy of 79.75\% was achieved. \\
\hline
Wimmer et al. (2008) \cite{wimmer2008facial} & face & anger, disgust, fear, happy, sad, surprise & B21 robot & Binary Decision Tree & Experimental evaluation reports a recognition rate of 70\% on the Cohn–Kanade facial expression database, and 67\% in a robot scenario. \\
\hline
Wu et al. (2019) \cite{wu2019weight} & face & angry, sad, disgust, fear, happy, neutral, surprise & - & Weight-Adapted Convolution Neural Network (WACNN) & The recognition accuracies of the proposed algorithm were higher than the deep CNN without HGA, indicating a better global optimization ability. \\
\hline
Yu and Tapus (2019) \cite{yu2019interactive} & face, body gesture & neutral, happy, angry, sad & Pepper & Random Forests (RF) & Developed a multimodal emotion recognition model with gait and thermal facial data, which is based on RF model and the modified confusion matrices of two individual models. \\
\end{longtable}
\end{center}
\end{landscape}
}
\twocolumn

In the machine learning algorithms, Viola Jones algorithm and OpenCV were popular choices for face detection. However, dlib face detector and ADABOOST algorithm were also used. To pre-process the images, greyscale conversion, image normalization, image augmentation (such as flip, zoom, rotate etc.) were usually applied. Further, some studies extract the important regions in faces like eyebrows, eyes, nose and mouth (also known as the acting units or AUs) that play an important role in FER. Others use local binary pattern (LBP) or histogram of oriented gradients (HOG) to extract the featural information. Finally, the classification is performed. Most of the studies perform classification for the six universally known emotions (happy, sad, disgust, anger, fear and surprise) and sometimes include a neutral expression. For final classification, k-Nearest Neighbor (KNN), Hidden Markov Model (HMM), Recurrent Neural Network (RNN), Convolutional Neural Network (CNN), Support Vector Machine (SVM) and Long Short-Term Memory (LSTM) are used.

In the deep learning algorithms, the input images are first pre-processed by performing face alignment, data augmentation and normalization. Then the images are directly fed into deep networks like CNN, RNN etc. which predict the emotion of the images. The most commonly used classification methods are explained in more detail below. They are arranged in the order in which they were invented.

KNN: Nearest neighbor based classifier was first invented in the 1950s \cite{fix1951discriminatory}. In KNN \cite{4406010}, given the training instances and the class labels, the class label of an unknown instance is predicted. KNN is based on a distance function that measures the difference between two instances. While the Euclidean distance formula is mostly used, there are also other distance formulae such as Hamming distance which can be used.

HMM: An HMM \cite{18626} was introduced in the late 1960s. It is a doubly embedded stochastic process, bearing a hidden stochastic process (a Markov chain) that is only visible through another stochastic process, producing a sequence of observations. The state sequence can be learned using Viterbi algorithm or Expectation-Modification (EM) algorithm. 

RNN: RNN \cite{lipton2015critical} was introduced in the 1980s. RNN is a feed-forward neural network that has an edge over adjacent time steps, introducing a notion of time. Hence, RNN is mainly used for a dynamic data input that has a temporal sequence. In RNN, a state depends upon the current input as well as the state of the network at the previous time step, making it possible to contain information from a long time window. 

CNN: Convolutional Networks \cite{5537907} were invented in 1989 \cite{lecun1989backpropagation}. CNNs are trainable multistage architectures composed of multiple stages. The input and output of each stage are sets of arrays called feature maps. Each stage of CNN is composed of three layers- a filter bank layer, a non-linearity layer and a feature pooling layer. The network is trained using the backpropagation method. They are used for end-to-end recognition wherein given the input image, the output is predicted by CNNs. They are even used as feature extractors which are further connected with neural networks layers like LSTM or RNN for the prediction.

SVM: SVM was invented by Vapnik in 1995 \cite{vapnik95}. In SVM \cite{788640}, the training data can be separated by a hyperplane.

LSTM: LSTM \cite{gers2000recurrent} was invented by Hochreiter and Schmidhuber in 1997 \cite{hochreiter1997long}. It also has recurrent connections but unlike RNN, it is capable of learning long-term dependencies. 

Table \ref{tab4} summarizes the major purpose, application areas, advantages, disadvantages and frequency of use for commonly used algorithms. For the frequency of use, only the number of papers that implement facial expression recognition during HRI or in real-time scenarios were counted. Although RNN has not been used for facial expression recognition during HRI or in real-time, some studies perform facial expression recognition on predefined datasets using RNN. 

27 studies on facial expression recognition during HRI were reviewed. Some of the studies have not been performed on a robot platform. These studies perform emotion recognition in real-time and mention HRI as their intended application. The studies are summarized in Table \ref{tab6}. Here, studies that perform facial expression recognition on predefined datasets or studies that perform facial expression recognition but not in real-time were not included.

\begin{table*}
{
\caption{Studies on FER; Note: Studies listed according to accuracy level}\label{tab0}
\subcaption{With accuracy greater than 90\% on static input i.e., human images}\label{tab1}

\begin{tabular}{p{5.0cm} p{4.00cm} p{2cm} p{1.5cm} p{3cm}}
\hline
\textbf{Study} & \textbf{Dataset} & \textbf{Algorithm} & \textbf{Classes} & \textbf{Accuracy} \\
\hline
Mistry et al. (2017) \cite{7456259} & CK+/MMI & SVM & 7 & 100\%/94.66\% \\
Kotsia and Pitas (2007) \cite{4032815} & CK & SVM & 6 & 99.7\%  \\
Hossain et al. (2017) \cite{7862118} & Jaffe/CK & GMM & 7 & 99.8\%/99.7\%  \\
Kar et al. (2017) \cite{10.1007/978-981-10-2107-7-19} & CK+ & BPNN & 6 & 99.51\% \\
Mliki et al. (2015) \cite{mliki} & CK/Jaffe & SVM & 7 & 99.24\%/96.50\% \\
Chen et al. (2017) \cite{7988558} & CK+/Jaffe & CNN & 7 & 99.1597\%/87.7350\%  \\
Zhang et al. (2016) \cite{zhang} & CK+ & CNN & 6 & 98.9\%  \\
Mayya et al. (2016) \cite{mayya2016automatic} & Jaffe/CK+ & CNN & 7 & 98.12\%/96.02\% \\
Minaee and Abdolrashidi (2019) \cite{minaee2019deep} & CK+/Jaffe & CNN & 7 & 98.0\%/92.8\% \\
Nwosu et al. (2017) \cite{8328557} & Jaffe/CK+ & CNN & 7 & 97.71\%/95.72\% \\
Yang et al. (2017) \cite{8273654} & CK+/Oulu-Casia/Jaffe & CNN & 6 & 97.02\%/92.89\%/92.21\% \\
Yang et al. (2018) \cite{8214102} & CK+/Oulu-Casia/Jaffe & WMDNN & 6 & 97.0\%/92.3\%/92.2\%   \\
Ding et al. (2017) \cite{7961731} & CK+/Oulu-Casia & CNN & 8/6 & 96.8\%/87.71\% \\
Gogi{\'c} et al. (2018) \cite{gogic2018fast} & CK+/Jaffe/ MMI & NN & 7 & 96.48\%/85.88\%/73.73\%  \\
Kim et al. (2019) \cite{8673885} & CK+/Jaffe & CNN & 6 & 96.46\%/91.27\%  \\
Hua et al. (2019) \cite{8643924} & Jaffe & CNN & 7 & 96.44\% \\
Mannan et al. (2015) \cite{10.1007/978-3-319-22053-6-33} & CK+ & SVM & 7 & 96.36\% \\
Ruiz-Garcia et al. (2018) \cite{ruiz-garcia2018} & KDEF/CK+ & CNN-SVM & 7 & 96.26\%/95.87\%  \\
Hamester et al. (2015) \cite{hamester} & Jaffe & CNN & 7 & 95.8\% \\
Meng et al. (2017) \cite{7961791} & CK+/MMI & CNN & 6 & 95.27\%/71.55\%  \\
Liliana et al. (2017) \cite{10.1145/3127942.3127958} & CK+ & SVM & 7 & 93.93\% \\
Ferreira et al. (2018) \cite{8472816} & CK+/Jaffe & CNN & 8/6 & 93.64\%/89.01\% \\
Mollahosseini et al. (2016) \cite{7477450} & CK+ & DNN & 7 & 93.2\% \\
Yaddadenet al. (2016) \cite{10.1145/2910674.2910703} & Jaffe/KDEF & KNN & 7 & 92.29\%/79.69\% \\
\end{tabular}
\subcaption{With accuracy greater than 90\% on dynamic input i.e., human videos}\label{tab2}
\begin{tabular}{p{5cm} p{4.00cm} p{2cm} p{1.5cm} p{3cm}}
\hline
\textbf{Study} & \textbf{Dataset} & \textbf{Algorithm} & \textbf{Classes} & \textbf{Accuracy} \\
\hline
Liang et al. (2020) \cite{liang2020deep} & CK+/Oulu-Casia/MMI & CNN-BiLSTM & 6 & 99.6\%/91.07\%/80.71\% \\
Carcagn{\`\i} et al. (2015) \cite{carcagni2015facial} & CK+ & SVM & 7 & 98.5\% \\
Wu et al. (2015) \cite{7163116} & CK+ & HMM & 7 & 98.54\% \\
Zhang et al. (2017) \cite{7890464} & CK+/Oulu-Casia/MMI & CNN-RNN & 6 & 98.5\%/86.25\%/81.18\% \\
Kotsia et al. (2007) \cite{4217476} & CK & SVM & 6 & 98.2\%  \\
Uddin et al. (2017) \cite{7867858} & Depth & DBN & 6 & 96.67\% \\
Zhao et al. (2017) \cite{zhao2017facial} & CK+/Oulu-Casia/MMI & SVM & 7 & 95.8\%/74.37\%/71.92\%  \\
Elaiwat et al. (2016) \cite{elaiwat2016spatio} & CK+/MMI & RBM & 7 & 95.66\%/81.63\%  \\
Uddin et al. (2017) \cite{8119492} & CK & CNN & 6 & 95.42\%  \\
Sikka et al. (2015) \cite{sikka2015exemplar} & CK+/Oulu-Csia & HMM & 7 & 94.60\%/75.62\% \\
Kabir et al. (2017) \cite{7929271} & Depth & HMM & 6 & 94.17\%  \\
\end{tabular}
\subcaption{FER in real-time i.e., on dynamic input during HRI}\label{tab3}
\begin{tabular}{p{5.0cm} p{2.0cm} p{2.0cm} p{1.5cm} p{1.5cm} p{3cm}}
\hline
\textbf{Study} & \textbf{Robot} & \textbf{Sensor} & \textbf{Algorithm} & \textbf{Classes} & \textbf{Accuracy} \\
\hline
Cid et al. (2014) \cite{cid2014} & Muecas & camera & Bayesian & 5 & 94\%  \\
Meghdari et al. (2016) \cite{meghdari2016} & Alice & Kinect & ANN & 6 & 92.52\%  \\
Simul et al. (2016) \cite{7860248} & Ribo & webcam & SVM & 5 & 86\%  \\
Bera et al. (2019) \cite{Bera2019TheEI} & Pepper & camera & CNN & 4 & 85.33\%  \\
Liu et al. (2017) \cite{liu2017} & mobile robot & Kinect & ELM & 7 & above 80\%  \\
Yu and Tapus (2019) \cite{yu2019interactive} & Pepper & camera & RF & 4 & 78.125\% \\ 
Webb et al. (2020) \cite{9207494} & Nao & camera & CNN & 8 & 79.75\%  \\
Chen et al. (2018) \cite{8630711} & XiaoBao & camera & CNN & 7 & 79\%  \\
Barros et al. (2015) \cite{7363421} & iCub & RGB camera & CNN & 3 & 74.2\%  \\
Ruiz-Garcia et al. (2018) \cite{ruiz-garcia2018} & Nao & built-in camera & CNN-SVM & 7 & 68.75\%  \\
Wimmer et al. (2008) \cite{wimmer2008facial} & B21 robot & camera & Binary Decision Tree & 6 & 67\% \\
\hline
\end{tabular}
}
{
\\
\emph{Notes}: Cohn-Kanade (CK), Japanese Female Facial Expression (Jaffe), Generative Adversarial Network (GAN), k-Nearest Neighbor (KNN), Hidden Markov Model (HMM), Recurrent Neural Network (RNN), Convolutional Neural Network (CNN), Support Vector Machine (SVM), Long Short-Term Memory (LSTM), Weighted Mixture Deep Neural Network (WMDNN), Neural Network (NN), Artificial Neural Network (ANN), Extreme Learning Machine (ELM), Back Propagation Neural Network (BPNN), Deep Belief Network (DBN), Random Forests (RF)}
\end{table*}

\subsection{FER on predefined dataset}
\label{sec:31}
\begin{figure*}[ht]
\includegraphics[width=\textwidth]{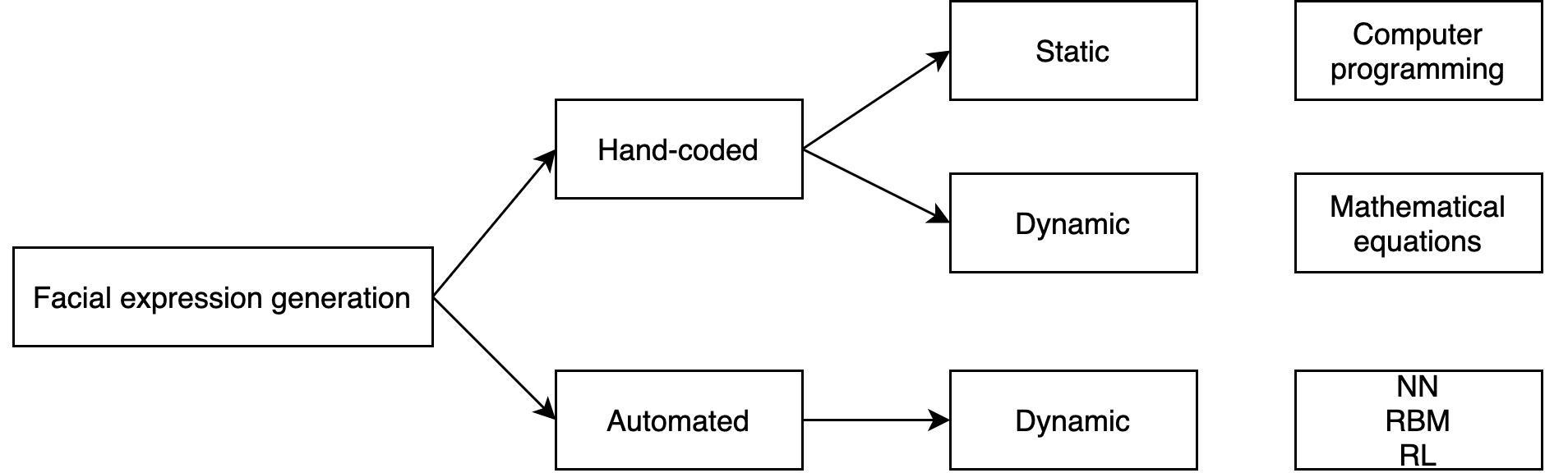} 
\emph{Notes}: Neural Network (NN), Restricted Boltzmann Machine (RBM), Reinforcement Learning (RL) 
\caption{Facial expression generation techniques}
\label{fig4}
\end{figure*}

Although the goal of this study is to perform FER in real-time and during HRI, the studies on real-time FER are compared with FER on predefined datasets. FER has been carried out on static human images as well as on dynamic human video clips. While some studies, perform facial recognition on still images, others perform facial recognition on videos. In Datcu and Rothkrantz \cite{datcu}, they show that there is an advantage in using data from video frames over still images. This is because videos contain temporal information that is absent in still images.

Results of studies with above 90\% accuracy in FER on still images are summarized in Table \ref{tab1} and on videos are summarized in Table \ref{tab2}. Tables \ref{tab1} and \ref{tab2} are for comparison with Table \ref{tab3}. Studies are arranged according to their accuracy level. It should be noted that these studies are carried out on predefined datasets consisting of human images and videos and do not involve robots. There are a considerable number of studies that achieve accuracy greater than 90\% on CK+, Jaffe and Oulu-Casia datasets on both still images and videos.

\subsection{FER in real-time}
It is easier to achieve high accuracy while performing emotion recognition on predefined datasets as they are recorded under controlled environmental conditions. On the other hand, it is difficult to achieve the same level of accuracy when performing emotion recognition in real-time when the movements are spontaneous. It should be noted that studies that perform facial expression recognition in real-time were carried out under controlled laboratory conditions with little variation in lighting conditions and head poses.

As this study is about facial expressions in HRI, for a robot to be able to recognize emotion, emotion recognition has to be performed in real-time. Table \ref{tab3} provides studies with facial expression recognition in real-time for HRI. Here, the accuracies are comparatively lower than the accuracies for predefined datasets. As can be seen in Table \ref{tab3}, only two studies have an accuracy greater than 90\%. The robots that are used in the studies are either robotic heads or humanoid robots such as Pepper, Nao, iCub etc. Many studies that perform facial expression recognition in real-time use CNNs, making it a popular choice for facial expression recognition \cite{Bera2019TheEI,9207494,8630711,7363421,7363421}. However, the highest accuracy is achieved by Bayesian and Artificial Neural Network (ANN) methods for facial expression recognition in real-time.

\section{Facial emotion expression by robots}
\label{sec:4}
For robots to be empathic, it is necessary that the robots not only be able to recognize human emotions but also be able to generate emotions using facial expressions. Several studies enable robots to generate facial expressions either in a hand-coded or an automated manner (Figure \ref{fig4}). By hand-coded, we mean that the facial expressions are coded by moving the eyes and mouth of the robot in a desirous manner, and automated is when the emotions are learned automatically using machine learning techniques. 

16 studies on facial emotion expression in robots were reviewed. These studies are summarized in Table \ref{tab5}.

\subsection{Facial expression generation is hand-coded}
Earlier studies started by hand-coding the facial expressions in robots. There is a static as well as dynamic generation of facial expressions on robots.

Among the static methods, there is a humanoid social robot “Alice” that imitates human facial expressions in real-time \cite{meghdari2016}. Kim et al. \cite{kimjung} introduced an artificial facial expression imitation system using a robot head, Ulkni. As Ulkni is composed of 12 RC servos, with four Degrees of Freedom (DoFs) to control its gaze direction, two DoFs for its neck, and six DoFs for its eyelids and lips, it is capable of making the basic facial expressions after the position commands for actuators are sent from the PC. Bennett and Sabanovic \cite{Bennett2014DerivingMF} identified minimal features, i.e. movement of eyes, eyebrows, mouth and neck, which are sufficient to identify the facial expression.  
 
\onecolumn
\afterpage{
\begin{landscape}
\begin{center}
\newcolumntype{P}[1]{>{\RaggedRight\arraybackslash}p{##1}}
\begin{longtable}{|P{2cm}|P{1.75cm}|P{3.5cm}|P{1.5cm}|P{2.5cm}|P{6cm}|}
\caption{Detailed information about studies on emotion expression and Human Robot Interaction (HRI)} \label{tab5} \\

\hline \multicolumn{1}{|P{2cm}|}{\textbf{Authors (year)}} & \multicolumn{1}{P{1.75cm}|}{\textbf{Expression mode}} &
\multicolumn{1}{P{3.5cm}|}{\textbf{Expressed emotions}} &
\multicolumn{1}{P{1.5cm}|}{\textbf{Robot}} &
\multicolumn{1}{P{2.5cm}|}{\textbf{Coding}} &
\multicolumn{1}{P{6cm}|}{\textbf{Major findings}} \\ \hline 
\endfirsthead

\multicolumn{5}{c}%
{{\bfseries}} \\
\hline \multicolumn{1}{|P{2cm}|}{\textbf{Authors (year)}} & \multicolumn{1}{P{1.75cm}|}{\textbf{Expression mode}} &
\multicolumn{1}{P{3.5cm}|}{\textbf{Expressed emotions}} &
\multicolumn{1}{P{1.5cm}|}{\textbf{Robot}} &
\multicolumn{1}{P{2.5cm}|}{\textbf{Coding}} &
\multicolumn{1}{P{6cm}|}{\textbf{Major findings}} \\ \hline
\endhead

\endfoot

\hline 
\endlastfoot
Bennett and Sabanovic (2014) \cite{Bennett2014DerivingMF} & face & neutral, sadness, happiness, anger, fear, surprise & MiRAE & hand-coded & Identified minimal features, i.e. movement of eyes, eyebrows, mouth and neck are sufficient to identify the facial expression. \\
\hline
Breazeal (2003)  \cite{breazeal2003emotion} & face & anger, calm, disgust, fear, content, interest, sorrow, surprise, tired & Kismet & hand-coded & Kismet can generate emotions using an interpolation-based technique over a 3-D space where the three dimensions correspond to valence, arousal and stance. \\
\hline
Breazeal et al. (2005) \cite{brazeal} & face & & Leonardo & automated (Neural Network) & Built a robot capable of learning how to imitate facial expressions from simple imitative games played with a human, using biologically inspired mechanisms. \\
\hline
Churamani et al. (2018) \cite{churamani-barros} & face & anger, happiness, neutral, sadness, surprise & Nico & automated (RL) & Explored a continuous representation of expression on the NICO robot using the complete face LED matrix to generate expressions.\\
\hline
Cid et al. (2013) \cite{6696662} & face & sad, happy, fear, anger, neutral & Muecas & hand-coded & The output of the Bayesian classifier is imitated by the robotic head Muecas.\\
\hline
Esfandbod et al. (2019) \cite{9071837} & face & neutral, happy, sad, surprised, angry & RASA & hand-coded & From the subjects' viewpoint, the system's performance was fairly promising with a score of 4.1 out of 5. \\
\hline
Ge et al. (2008) \cite{ge2008} & face & happy, fear, surprise, disgust, sadness, anger & robotic head & hand-coded & The robot head are triggered to imitate human facial expressions by the emotion generator engine and can generate a vivid imitation according to the tester's facial expression. \\
\hline 
Horii et al. (2016) \cite{horii} & face, hands, voice & happiness, neutral, anger, sadness & iCub & automated (RBM) & The robot does not copy the human's expressions directly but generates expressions on its own after estimating the human's emotion. \\
\hline
Ilic et al. (2019) \cite{ilic2019} & face & anger, disgust, fear,
happiness, sadness, surprise, indifference & Aisoy & hand-coded & 
Learned a model of the emotional value of the robot’s facial expressions without humans’ explicit feedback. \\
\hline 
Kim et al. (2006) \cite{kimjung} & face & neutral, anger, happiness, fear, sadness, surprise & Ulkni & hand-coded & Introduced an artificial facial expression imitation system using a robot head. \\
\hline
Kishi et al. (2012) \cite{6386050} & face & anger, sadness, fear, disgust, surprise, happy & KOBIAN & hand-coded & Developed a new head for biped walking robot KOBIAN that could express the 6 basic emotions. \\
\hline
Liu et al. (2017) \cite{liu2017} & face & happy, angry, surprise, fear, disgust, sad, neutral & mobile robot & hand-coded & Generated facial expressions adapting to human emotions using a four-layer framework designed for the system to recognize human emotion based on HRI. \\
\hline
Maeda and Geshi (2018) \cite{maeda2018human} & face & anger, contempt, disgust, fear, happiness, neutral, sadness, surprise & TAPIA & hand-coded & An interactive communication method of a human and a robot based on the Markovian emotional model (MEM) by using the facial expression, was significantly better than using an identical, symmetric or random emotion interaction methods. \\
\hline
Meghdari et al. (2016) \cite{meghdari2016} & face & happy, sad, angry, surprised, disgusted, neutral & Alice & hand-coded & A humanoid social robot “Alice” imitates human facial expressions in real-time. \\
\hline
Park et al. (2015) \cite{park} & face & anger, disgust, fear, happiness, sadness, surprise, dislike & robotic head & hand-coded & Made diverse facial expressions by changing their dynamics and increased the lifelikeness of a robot by adding secondary actions such as physiological movements (eye blinking and sinusoidal motions concerning respiration). \\
\hline 
Yoo et al. (2011) \cite{6007468} & face & angry, disgust, fear, happiness, sadness, surprise & robotic head & hand-coded & A fuzzy integral-based generation method of composite facial expressions was proposed and demonstrated its effectiveness through the experiment with the developed robotic head. \\
\end{longtable}
\end{center}
\end{landscape}
}
\twocolumn

In this study, the main program called functions that specified facial expressions according to the direction (used to make or undo an expression) and degree (strength of the expression—i.e. smaller vs. larger). The facial expression functions would in turn call lower functions that moved specific facial components given a direction and degree, following the movement related to specific AUs in the facial acting coding system (FACS).

Breazeal's \cite{breazeal2003emotion} robot Kismet generated emotions using an interpolation-based technique over a 3-D space, where the three dimensions correspond to valence, arousal and stance. The expressions become intense as the affect state moves to extreme values in the affect space. Park et al. \cite{park} made diverse facial expressions by changing their dynamics and increased the lifelikeness of a robot by adding secondary actions such as physiological movements (eye blinking and sinusoidal motions concerning respiration). A second-order differential equation based on the linear affect-expressions space model is used to achieve the dynamic motion for expressions. Prajapati et al. \cite{prajpati13} used a dynamic emotion generation model to convert the facial expressions derived from the human face into a more natural form before rendering them on the robotic face. The model is provided with the facial expression of the person interacting with the system and corresponding synthetic emotions generated are fed to the robotic face. 

\emph{Summary of findings}: The robot faces are capable of making basic facial expressions as they contain enough DoFs in the eyes and mouth. They are able to generate static emotions \cite{meghdari2016,kimjung,Bennett2014DerivingMF}. Additionally, the robot faces are able to generate dynamic emotions \cite{breazeal2003emotion,park,prajpati13}.

\subsection{Facial expression generation is automated}
Some of the studies automatically generate facial expressions on robots. Unlike hand-coded techniques where the commands for the position of features like eyes and mouth are sent from the computer, here, the facial expressions are generated using machine learning techniques such as neural networks and RL.

Breazeal et al. \cite{brazeal} presented a robot Leonardo that can imitate human facial expressions. They use neural networks to learn the direct mapping of a human's facial expressions onto Leonardo's own joint space. In Horii et al. \cite{horii}, the robot does not directly imitate the human but estimates the correct emotion and generates the estimated emotion using RBM. RBM \cite{hinton2006reducing} is a generative model that represents the generative process of data distribution and latent representation, and can generate data from latent signals \cite{sukhbaatar2011robust,ngiam2011multimodal,NIPS2012-4683}.

Li and Hashimoto \cite{Li-hashimoto} developed a KANSEI communication system based on emotional synchronization. KANSEI is a Japanese term that means emotions, feeling, sensitivity etc. The KANSEI communication system first recognizes human emotion and maps the recognized emotion to the emotion generation space. Finally, the robot expresses its emotion synchronized with the human's emotion in the emotion generation space. When the human changes his/her emotion, the robot also synchronizes its emotion with the human's emotion, establishing a continuous communication between the human and the robot. It was found that the subjects became more comfortable with the robot and communicated more with the robot when there was emotional synchronization.

In Churamani et al. \cite{churamani-barros}, the robot Nico learned the correct combination of eyebrow and mouth wavelet parameters to express its mood using RL. The learned expressions looked slightly distorted but were sufficient to distinguish between various expressions. The robot could also generate expressions that were not limited to the basic five expressions that were learned. For a mixed emotional state (for example, anger mixed with sadness), the model was able to generate novel expression representations representing the mixed state of the mood.

\emph{Summary of findings}. In all of the above studies, the robots learn to generate facial expressions automatically using machine learning techniques. While Breazeal \cite{brazeal}, Li and Hashimoto \cite{Li-hashimoto} used direct mapping of human facial expressions, Horii et al. \cite{horii} generated the estimated human's emotion on the robot. In Churamani et al. \cite{churamani-barros}, the robot was able to associate the learned expressions with the context of the conversation.

\begin{table*}[ht]
{
\caption{Possible categories for facial recognition in the wild}
\label{tab7}
\begin{tabular}{p{3cm} p{3cm} p{5cm} p{2cm} p{2cm}}
\hline
\textbf{Category} & \textbf{Description} & \textbf{Examples} & \textbf{Related studies} & \textbf{Applied algorithm} \\
\hline
\rowcolor{Gray}
Basic emotional facial expressions & Expression of basic emotions, such as FACS by Ekman (2001) & Happiness, sadness, anger, disgust, fear, surprise, neutral & \cite{9039580,buciu2005facial,Rouast2019DeepLF,7990582,7374704,6940284,gunes2013categorical,5771357,canedo} & KNN, HMM, RNN, CNN, SVM and LSTM \\
\rowcolor{LightGray}
Face movements & Movements of the face itself (as a whole) & Moving forward, backward, turning around, turning sideways jerking the head forward, spinning & \cite{6977018,8995782,8578452,8967874} & GAN,CNN \\
\rowcolor{LightGray}
Situation-specific face occlusions & Occlusions of the face due to situational requirements & Mouth-nose mask, glasses, hand in front of face, resting hand on mouth, headset & \cite{Yong18,8545853,8974606} & CNN\\
\rowcolor{LightGray}
Permanent face features & Permanently installed features of the face & Artificial eye, beard & \cite{Yong18,8545853,8974606} & CNN \\
Situation-specific expressions & Facial expressions of the face due to situational requirements & Nodding, blinking, looking down, yawning, shake head in agreement, eye roll, closing eyes, lip biting, pursed lips, sticking tongue out, winking & & \\
Side activities during facial expression & Additional activities during the expressions of emotions & Talking, eating, drinking, brushing teeth, fixing hair, combing hair, biting nails, cleaning eyes with a hand, coughing, supporting the face with a hand, itching on face, blowing nose, sneezing, rubbing eyes, sipping through a straw, applying cream & & \\
\end{tabular}
}
Notes: 1) examples were generated based on 50 life observations and 50 video-based observations by the authors; 2) dark grey = very well understood in research; medium grey = partly understood in research; white = hardly understood
\end{table*}

\section{Discussion}
\label{sec:5}
\subsection{Summary of the state of the art}
There are already studies having high accuracy (greater than 90\%) in facial expression recognition on CK+, Jaffe and Oulu-Casia datasets. (see Tables \ref{tab1} and \ref{tab2}). The accuracies on CK+, Jaffe and Oulu-Casia datasets have been as high as 100\%, 99.8\% and 92.89\% respectively. In comparison to this, the accuracy for facial expression recognition in real-time is not as high.

Zhang et al. \cite{zhang} used a deep convolutional network (DCN) that had an accuracy of 98.9\% on CK+ dataset and 55.27\% on Static Facial Expressions in the Wild (SFEW) dataset. Here, the same network produced very different results for two different datasets. SFEW \cite{6130508} consists of close to a real-world environment extracted from movies. The database covers unconstrained facial expressions, varied head poses, large age range, occlusions, varied focus, different resolution of faces, and close to real-world illumination. In Zhang et al. \cite{zhang} the accuracy for "in the wild" settings was considerably lower than on CK+ dataset, implying that the expression recognition algorithms can still not handle the variations in environment, head poses etc. in real-world settings. 

Table \ref{tab7} provides possible categories for facial recognition in the wild. It contains the basic emotional facial expressions, situation-specific face occlusions, permanent face features, face movements, situation-specific expressions and side activities during facial expressions. 

Most of the current research in facial expression recognition relates to the first category of basic emotional facial expression. Survey articles on facial expression recognition have been cited in the table \ref{tab7} \cite{9039580,buciu2005facial,Rouast2019DeepLF,7990582,7374704,6940284,gunes2013categorical,5771357,canedo}. For more details on individual studies, refer to Table \ref{tab0}. Facial expression recognition in the presence of situation-specific face occlusions like a mouth-nose mask, glasses, hand in front of face etc. has also been studied \cite{Yong18,8545853,8974606}. Pose invariant facial expression recognition when the face is moving or turned sideways has also been partially studied \cite{6977018,8995782,8578452,8967874}.

For the facial expression generation, robots can make certain basic facial expressions by moving their eyes, mouth and neck. However, they cannot make as many expressions as human beings due to the limited number of DoFs present in a robot's face. There are relatively fewer studies for automated facial expression generation in robots \cite{brazeal,horii,Li-hashimoto,churamani-barros}. While the robots are capable of displaying their facial expressions by manually coding the movement of the eyes and mouth, there are fewer studies that would make a robot learn to display its facial expressions automatically.

Most of the studies on facial expression generation have been carried out on robotic heads or humanoid robots like iCub and Nico \cite[e.g.][]{breazeal2003emotion,brazeal,horii,churamani-barros}. In Becker-Asano and Ishiguro \cite{becker2011evaluating}, Geminoid F's facial actuators are tuned such that the readability of its facial expressions is comparable to a real person's static display of emotional expression. It was found that the android's emotional expressions were more ambiguous than that of a real person and 'fear' was often confused with 'surprise'.

An advantage of automated facial expression generation over hand-coded facial expression generation is that in automated facial expression generation, a robot could learn mixed expressions than simply the learned expressions. Unlike in hand-coded facial expression generation, where a robot can only express the emotions that it has learned, in Churamani et al. \cite{churamani-barros}, the robot could express complex emotions that were made up of a combination of emotions.

\subsection{Future research}
Although facial expression recognition under specific settings has high accuracy and robots can express basic emotions through facial expressions, there are several possible directions for future research in this area. 

\textit{Suggestion 1: Performing facial expression recognition in the wild needs to be emphasized upon.} \\
To efficiently recognize facial expressions in real-time and in a real-world environment, the robot should be able to perform facial expression recognition with varied head poses, varied focus, presence of occlusions, different resolutions of the face and varied illumination conditions. The studies that perform facial expression recognition in real-time are limited to a laboratory environment which is far different from a real-world scenario. A good study would be the one where facial expression recognition in the wild is performed.

Some studies perform facial expression recognition in the wild, but their accuracy is much less than the accuracy on predefined datasets like CK+, Jaffe, MMI etc. To increase the efficiency of facial expression recognition in real-world scenarios, the performance of facial expression recognition in the wild needs to be improved. This can also be used to recognize facial expressions in real-time. Based on this, a direct adaptation of emotions would make HRI smoother. 

\textit{Suggestion 2: Facial expressions during activities like talking, nodding etc. need to be studied.} \\
Situation-specific expressions (nodding, yawning, blinking, looking down) and side activities during facial expressions (talking, eating, drinking, sneezing) in Table \ref{tab7} have not been studied. To understand vivid expressions, it is required to be able to recognize facial expressions for all categories. Humans also express emotions while interacting with someone verbally, such as smiling while speaking when they are happy. In this case, it should be possible to recognize a smile during speech.

\textit{Suggestion 3: Combine facial expression recognition with the data from other modalities such as voice, text, body gestures and physiological data to improve the emotion recognition rate.} \\
Although this overview focuses on facial expression recognition, it may be possible to control one's face and not express the emotion one is truly experiencing. Some studies combine facial expression recognition with audio data, body gestures or physiological data for an improved emotion recognition \cite{ma2020learning,gunes2007bi,huang2017fusion}. There are very few studies that combine facial data with both audio and physiological data \cite{ringeval2015avec,ringeval201522} and studies that analyze all modalities (face, voice, text, body gestures and physiological signals) have not been found. Humans can recognize the emotion of a person quickly and effectively by taking into account their facial expression, body gestures, voice and words. Combining facial, audio, text and body gestures with physiological data could lead to a higher emotion recognition rate by machine learning algorithms than by humans.

\textit{Suggestion 4: How should a robot react towards a given human emotion?} \\
In HHI, a human's reaction to a given emotion is either a result of parallel empathy or reactive empathy \cite{Davis2018}. It should be studied with which emotion should a robot appropriately react to a given human emotion. Moreover, it needs to be studied if a robot should be able to express negative emotions. Most of the existing studies allow a robot to be able to express basic emotions (anger, fear, happiness, neutral, sadness, surprise). It may be reasonable for a robot to react with a sad expression when a human being expresses anger. But, should a robot be able to express extreme emotions such as anger?

For facial expression generation, while robots are capable of displaying facial expressions both static and dynamic, they are unable to generate facial expressions when they are speaking. For example, robots could smile while talking to express their happiness or they could speak with a frown when angry. Robots could also express their emotions through partial facial or bodily gestures instead of showing a full face expression. For example, tilting head down to express sadness, frowning to express anger, eyes wide open to express surprise and raising eyebrows. 

\textit{Suggestion 5: Robots should be able to recognize and generate facial expressions with various intensities.} \\
Emotions form a continuous range and can have various intensities. If one is less happy, one would smile less. Similarly, if someone is very happy, the smile would also be big. It should be possible to recognize not just the emotion but the intensity of emotion.
Moreover, in most of the existing studies, robots express their emotions with only one configuration per emotion. Robots should also be able to express their emotions with different intensities. Finally, it needs to be studied whether the intensity of emotion with which a robot reacts to a given human emotion has any effects on the human and whether these effects are positive or negative.  

\textit{Suggestion 6: Robots should be able to express their emotions through a combination of body gestures and facial expressions.} \\
While in this overview, we focus on robotic facial expressions, there are other articles where emotional expression is performed through the robot's body postures \cite{beck2013interpretation,mccoll2014recognizing,costa2013facial,marmpena2019generating,inthiam2019mood,cohen2010recognizing}. A potential future study could be to compare the robot's facial expressions with robot's bodily expressions and also with the combination of facial and bodily expressions to see if there is any difference in the recognition of these.

\textit{Suggestion 7: Robots should be able to both recognize and generate complex emotions such as that of thinking, calm and bored states.} \\
For both facial expression recognition and generation, there is a need to go beyond the basic seven emotions to recognizing and generating more complex emotions such as calm, fatigued, bored etc. It might be difficult to generate complex emotions given the hardware limitations of the robot, but if this is made possible, robots could express a wider range of emotions similar to human beings. 

\section{Conclusion}
\label{sec:6}
This overview emphasizes the recognition of human facial expressions and the generation of robotic facial expressions. There are already plenty of studies having high accuracy for facial expression recognition on pre-existing datasets. Accuracy on facial expression recognition in the wild is considerably lower than the experiments which have been conducted under controlled laboratory conditions. For human facial emotion recognition, future work would be to improve emotion recognition for non-frontal head poses in presence of occlusions (i.e. emotion recognition in the wild). It should be made possible to recognize emotions during speech as well emotions with varying intensities. In the case of facial expression generation in robots, robots are capable of making the basic facial expressions. Few studies perform autonomous facial generation in robots. In the future, there could be studies comparing robotic facial expressions with the robot's bodily expressions and also with a combination of facial and bodily expressions to see if there is any difference in recognizing these. Robots should be able to express their emotion with partial bodily or facial gestures while speaking. They should also be express their emotions with various intensities instead of a single configuration per emotion. Lastly, there is a need to go beyond the basic seven expressions for both facial expression recognition and generation.  

\begin{acknowledgements}
The authors thank Vignesh Prasad for his insightful comments. 
\end{acknowledgements}
\section*{Compliance}
Compliance with Ethical Standards: The authors declare that there are no compliance issues with this research. Funding: This research was funded by the German Research Foundation (DFG, Deutsche Forschungsgemeinschaft). The authors also thank the ZEVEDI Hessen and the leap in time foundation for the grateful funding of the project. Conflict of Interest: The authors declare that they have no conflict of interest.
%
%

\bibliographystyle{spmpsci}      
\bibliography{biblio}   

\end{document}